\pdfoutput=1

\documentclass[11pt]{article}

\usepackage[final]{acl}

\usepackage{times}
\usepackage{latexsym}

\usepackage{booktabs}
\usepackage{multirow}
\usepackage{graphicx}
\usepackage[table]{colortbl}

\usepackage[T1]{fontenc}

\usepackage[utf8]{inputenc}

\usepackage{microtype}

\title{BERTopic: Neural topic modeling with a class-based TF-IDF procedure}

\author{Maarten Grootendorst \\
  \texttt{maartengrootendorst@gmail.com}}

\begin{document}
\maketitle
\begin{abstract}

Topic models can be useful tools to discover latent topics in collections of documents. Recent studies have shown the feasibility of approach topic modeling as a clustering task. We present BERTopic, a topic model that extends this process by extracting coherent topic representation through the development of a class-based variation of TF-IDF. More specifically, BERTopic generates document embedding with pre-trained transformer-based language models, clusters these embeddings, and finally, generates topic representations with the class-based TF-IDF procedure. BERTopic generates coherent topics and remains competitive across a variety of benchmarks involving classical models and those that follow the more recent clustering approach of topic modeling.

\end{abstract}

\section{Introduction}

To uncover common themes and the underlying narrative in text, topic models have proven to be a powerful unsupervised tool. Conventional models, such as Latent Dirichlet Allocation (LDA) \citep{blei2003latent} and Non-Negative Matrix Factorization (NMF) \citep{fevotte2011algorithms}, describe a document as a bag-of-words and model each document as a mixture of latent topics. 

One limitation of these models is that through bag-of-words representations, they disregard semantic relationships among words. As these representations do not account for the context of words in a sentence, the bag-of-words input may fail to accurately represent documents. 

As an answer to this issue, text embedding techniques have rapidly become popular in the natural language processing field. More specifically, Bidirectional Encoder Representations from Transformers (BERT) \citep{devlin2018bert} and its variations \citep[e.g.,][]{lee2020biobert, liu2019roberta, lan2019albert}, have shown great results in generating contextual word- and sentence vector representations. The semantic properties of these vector representations allow the meaning of texts to be encoded in such a way that similar texts are close in vector space. 

Although embedding techniques have been used for a variety of tasks, ranging from classification to neural search engines, researchers have started to adopt these powerful contextual representations for topic modeling. \citet{sia2020tired} demonstrated the viability of clustering embeddings with centroid-based techniques, compared to conventional methods such as LDA, as a way to represent topics. From these clustered embeddings, topic representations were extracted by embedding words and finding those that are in close proximity to a cluster's centroid. Similarly, Top2Vec leverages Doc2Vec's word- and document representations to learn jointly embedded topic, document, and word vectors \cite{angelov2020top2vec, le2014distributed}. Comparable to \citet{sia2020tired}'s approach, documents are clustered and topic representations are created by finding words close to a cluster's centroid. Interestingly, although the topic representations are generated from a centroid-based perspective, the clusters are generated from a density-based perspective, namely by leveraging HDBSCAN \cite{mcinnes2017accelerated}. 

The aforementioned topic modeling techniques assume that words in close proximity to a cluster's centroid are most representative of that cluster, and thereby a topic. In practice, however, a cluster will not always lie within a sphere around a cluster centroid. As such, the assumption cannot hold for every cluster of documents, and the representation of those clusters, and thereby the topic might be misleading. Although \cite{sia2020tired} attempts to overcome this issue by re-ranking topic words based on their frequency in a cluster, the initial candidates are still generated from a centroid-based perspective. 

In this paper, we introduce BERTopic, a topic model that leverages clustering techniques and a class-based variation of TF-IDF to generate coherent topic representations. More specifically, we first create document embeddings using a pre-trained language model to obtain document-level information. Second, we first reduce the dimensionality of document embeddings before creating semantically similar clusters of documents that each represent a distinct topic. Third, to overcome the centroid-based perspective, we develop a class-based version of TF-IDF to extract the topic representation from each topic. These three independent steps allow for a flexible topic model that can be used in a variety of use-cases, such as dynamic topic modeling.

\section{Related Work}
\label{sec:related} 
In recent years, neural topic models have increasingly shown success in leveraging neural networks to improve upon existing topic modeling techniques \cite{terragni2021octis, cao2015novel, zhao2021topic, larochelle2012neural}. The incorporation of word embeddings into classical models, such as LDA, demonstrated the viability of using these powerful representations \cite{liu2015topical, nguyen2015improving, shi2017we, qiang2017topic}. Foregoing incorporation into LDA-like models, there has been a recent surge of topic modeling techniques built primarily around embedding models illustrating the potential of embedding-based topic modeling techniques \cite{bianchi2020cross, dieng2020topic, thompson2020topic}. CTM, for example, demonstrates the advantage of relying on pre-trained language models, namely that future improvements in language models may translate into better topic models \cite{bianchi2020pre}.

Several approaches have started simplifying the topic building process by clustering word- and document embeddings \cite{sia2020tired, angelov2020top2vec}. This clustering approach allows for a flexible topic model as the generation of the clusters can be separated from the process of generating the topic representations. 

BERTopic builds on top of the clustering embeddings approach and extends it by incorporating a class-based variant of TF-IDF for creating topic representations.

\section{BERTopic}
\label{sec:methodology}

BERTopic generates topic representations through three steps. First, each document is converted to its embedding representation using a pre-trained language model. Then, before clustering these embeddings, the dimensionality of the resulting embeddings is reduced to optimize the clustering process. Lastly, from the clusters of documents, topic representations are extracted using a custom class-based variation of TF-IDF. 

\subsection{Document embeddings}
\label{sec:documentembeddings}
In BERTopic, we embed documents to create representations in vector space that can be compared semantically. We assume that documents containing the same topic are semantically similar. To perform the embedding step, BERTopic uses the Sentence-BERT (SBERT) framework \citep{reimers-2019-sentence-bert}. This framework allows users to convert sentences and paragraphs to dense vector representations using pre-trained language models. It achieves state-of-the-art performance on various sentence embedding tasks \cite{reimers2020making, thakur2020augmented}.

These embeddings, however, are primarily used to cluster semantically similar documents and not directly used in generating the topics. Any other embedding technique can be used for this purpose if the language model generating the document embeddings was fine-tuned on semantic similarity. As a result, the quality of clustering in BERTopic will increase as new and improved language models are developed. This allows BERTopic to continuously grow with the current state-of-the-art in embedding techniques. 

\subsection{Document clustering}
As data increases in dimensionality, distance to the nearest data point has been shown to approach the distance to the farthest data point \citep{aggarwal2001surprising, beyer1999nearest}. As a result, in high dimensional space, the concept of spatial locality becomes ill-defined and distance measures differ little. 

Although clustering approaches exist for overcoming this curse of dimensionality \cite{pandove2018systematic, steinbach2004challenges}, a more straightforward approach is to reduce the dimensionality of embeddings. Although PCA and t-SNE are well-known methods for reducing dimensionality, UMAP has shown to preservers more of the local and global features of high-dimensional data in lower projected dimensions \citep{2018arXivUMAP}. Moreover, since it has no computational restrictions on embedding dimensions, UMAP can be used across language models with differing dimensional space. Thus, we use UMAP to reduce the dimensionality of document embeddings generated in \ref{sec:documentembeddings} \citep{mcinnes2018umap-software}. 

The reduced embeddings are clustering used HDBSCAN \citep{mcinnes2017hdbscan}. It is an extension of DBSCAN that finds clusters of varying densities by converting DBSCAN into a hierarchical clustering algorithm. HDBSCAN models clusters using a soft-clustering approach allowing noise to be modeled as outliers. This prevents unrelated documents to be assigned to any cluster and is expected to improve topic representations.`Moreover, \cite{allaoui2020considerably} demonstrated that reducing high dimensional embeddings with UMAP can improve the performance of well-known clustering algorithms, such as k-Means and HDBSCAN, both in terms of clustering accuracy and time. 

\subsection{Topic Representation}
The topic representations are modeled based on the documents in each cluster where each cluster will be assigned one topic. For each topic, we want to know what makes one topic, based on its cluster-word distribution, different from another? For this purpose, we can modify TF-IDF, a measure for representing the importance of a word to a document, such that it allows for a representation of a term's importance to a topic instead.

The classic TF-IDF procedure combines two statistics, term frequency, and inverse document frequency \cite{joachims1996probabilistic}:

\begin{equation}
	W_{t,d} = tf_{t,d} \cdot \log({\frac{N}{df_t}})
\end{equation}

Where the term frequency models the frequency of term \textit{t} in document \textit{d}. The inverse document frequency measures how much information a term provides to a document and is calculated by taking the logarithm of the number of documents in a corpus \textit{N} divided by the total number of documents that contain \textit{t}. 

We generalize this procedure to clusters of documents. First, we treat all documents in a cluster as a single document by simply concatenating the documents. Then, TF-IDF is adjusted to account for this representation by translating documents to clusters:

\begin{equation}
	W_{t,c} = tf_{t,c} \cdot \log({1+\frac{A}{tf_t}})
\end{equation}

Where the term frequency models the frequency of term \textit{t} in a class \textit{c} or in this instance. Here, the class \textit{c} is the collection of documents concatenated into a single document for each cluster. Then, the inverse document frequency is replaced by the inverse class frequency to measure how much information a term provides to a class. It is calculated by taking the logarithm of the average number of words per class \textit{A} divided by the frequency of term \textit{t} across all classes. To output only positive values, we add one to the division within the logarithm. 

Thus, this class-based TF-IDF procedure models the importance of words in clusters instead of individual documents. This allows us to generate topic-word distributions for each cluster of documents. 

Finally, by iteratively merging the c-TF-IDF representations of the least common topic with its most similar one, we can reduce the number of topics to a user-specified value. 

\section{Dynamic Topic Modeling}
\label{sec:dtm}
Traditional topic modeling techniques are static in nature and do not allow for sequentially-organized of documents to be modeled. Dynamic topic modeling techniques, first introduced by \citep{blei2006dynamic} as an extension of LDA, overcome this by modeling how topics might have evolved over time and the extent to which topic representations reflect that. 

In BERTopic, we can model this behavior by leveraging the c-TF-IDF representations of topics. Here, we assume that the temporal nature of topics should not influence the creation of global topics. The same topic might appear across different times, albeit possibly represented differently. As an example, a global topic about cars might contain words such as "car" and "vehicle" regardless of the temporal nature of specific documents. Car-related documents created in 2020, however, might be better represented with words such as "Tesla" and "self-driving" whereas these words would likely not appear in car-related documents created in 1990. Although the same topic is assigned to car-related documents in 1990 and 2020, its representation might differ. Thus, we first generate a global representation of topics, regardless of their temporal nature, before developing a local representation. 

To do this, BERTopic is first fitted on the entire corpus as if there were no temporal aspects to the data in order to create a global view of topics. Then, we can create a local representation of each topic by simply multiplying the term frequency of documents at timestep \textit{i} with the pre-calculated global IDF values:

\begin{equation}
	W_{t,c,i} = tf_{t,c,i} \cdot \log({1+\frac{A}{tf_{t}}})
\end{equation}

A major advantage of using this technique is that these local representations can be created without the need to embed and cluster documents which allow for fast computation. Moreover, this method can also be used to model topic representations by other meta-data, such as author or journal. 

\subsection{Smoothing}
Although we can observe how topic representations are different from one time to another, the topic representation at timestep \textit{t} is independent of timestep \textit{t-1}. As a result, this dynamic representation of topics might not result in linearly evolving topics. When we expect linearly evolving topics, we assume that a topic representation at timestep \textit{t} depends on the topic representation at timestep \textit{t-1}. 

To overcome this, we can leverage the c-TF-IDF matrices that were created at each timestep to incorporate this linear assumption. For each topic and timestep, the c-TF-IDF vector is normalized by dividing the vector with the L1-norm. When comparing vectors, this normalization procedure prevents topic representations from having disproportionate effects as a result of the size of the documents that make up the topic. 

Then, for each topic and representation at timestep \textit{t}, we simply take the average of the normalized c-TF-IDF vectors at \textit{t} and \textit{t-1}. This allows us to influence the topic representation at \textit{t} by incorporating the representation at \textit{t-1}. Thus, the resulting topic representations are smoothed based on their temporal position. 

It should be noted that although we might expect linearly evolving topics, this is not always the case. Hence, this smoothing technique is optional when using BERTopic and will be reflected in the experimental setup.

\begin{table*}[ht]
    \centering
    \begin{tabular}{ccccccccc}
        \toprule
        {} & \multicolumn{2}{c}{\textbf{20 NewsGroups}} & \multicolumn{2}{c}{\textbf{BBC News}} 
        & \multicolumn{2}{c}{\textbf{Trump}}\\
        \cmidrule(lr){2-3} \cmidrule(lr){4-5} \cmidrule(lr){6-7} \cmidrule(lr){8-9}
        {} & TC & TD & TC & TD & TC & TD \\
        \midrule
        LDA & .058 & .749 & .014 & .577 &  \cellcolor{green!10}-.011\hspace{.11cm} & .502\\
        NMF & .089 & .663 & .012 & .549 & \cellcolor{green!40}.009 & .379\\
        T2V-\textit{MPNET} & .068 & .718 & -.027\hspace{.11cm} & .540 & -.213\hspace{.11cm} & \cellcolor{green!40}.698\\
        T2V-\textit{Doc2Vec} &  \cellcolor{green!80}.192 & \cellcolor{green!10}.823 & \cellcolor{green!80}.171 & \cellcolor{green!10}.792 & -.169\hspace{.11cm} & .658\\
        CTM & \cellcolor{green!10}.096 & \cellcolor{green!80}.886 & \cellcolor{green!10}.094 & \cellcolor{green!80}.819 & \cellcolor{green!40}.009 & \cellcolor{green!80}.855\\
        BERTopic-\textit{MPNET} & \cellcolor{green!40}.166 & \cellcolor{green!40}.851 & \cellcolor{green!40}.167 & \cellcolor{green!40}.794 & \cellcolor{green!80}.066 & \cellcolor{green!10}.663\\
        \bottomrule
    \end{tabular}
    \caption{Ranging from 10 to 50 topics with steps of 10, topic coherence (TC) and topic diversity (TD) were calculated at each step for each topic model. All results were averaged across 3 runs for each step. Thus, each score is the average of 15 separate runs.}
    \label{tab:main-results}
\end{table*}

\section{Experimental Setup}
\label{sec:experiment}

OCTIS (Optimizing and Comparing Topic models is Simple), an open-source python package, was used to run the experiments, validate results, and preprocess the data \cite{terragni2021octis}.

Both the implementation of BERTopic as well as the experimental setup are freely available online. \footnote{\url{https://github.com/MaartenGr/BERTopic}}\footnote{\url{https://github.com/MaartenGr/BERTopic_evaluation}}

\subsection{Datasets}
Three datasets were used to validate BERTopic, namely 20 NewsGroups, BBC News, and Trump's tweets. We choose to thoroughly preprocess the 20 NewsGroups and BBC News datasets, and only slightly preprocess Trump's tweets to generate more diversity between datasets. 

The 20 NewsGroups dataset\footnote{\url{https://github.com/MIND-Lab/OCTIS/tree/master/preprocessed_datasets/20NewsGroup}} contains 16309 news articles across 20 categories \cite{lang1995newsweeder}. The BBC News dataset\footnote{\url{https://github.com/MIND-Lab/OCTIS/tree/master/preprocessed_datasets/BBC_news}} contains 2225 documents from the BBC News website between 2004 and 2005 \cite{greene06icml}. Both datasets were retrieved using OCTIS, and preprocessed by removing punctuation, lemmatization, removing stopwords, and removing documents with less than 5 words. 

To represent more recent data in a short-text form, we collected all tweets of Trump\footnote{\url{https://www.thetrumparchive.com/faq}} before and during his presidency. The data contains 44253 tweets, excluding re-tweets, between 2009 and 2021. In both datasets, we lowercased all tokens. 

To evaluate BERTopic in a dynamic topic modeling setting, Trump's tweets were selected as they inherently had a temporal nature to them. Additionally, the transcriptions of the United Nations (UN) general debates between 2006 and 2015\footnote{\url{https://runestone.academy/runestone/books/published/httlads/_static/un-general-debates.csv}} were analyzed \citep{baturo2017understanding}. The Trump dataset was binned to 10 timesteps and the UN datasets to 9 timesteps. 

\subsection{Models}
BERTopic will be compared to LDA, NMF, CTM, and Top2Vec. LDA and NMF were run through OCTIS with default parameters. The \textit{"all-mpnet-base-v2"} SBERT model was used as the embedding model for BERTopic and CTM \citep{song2020mpnet}. Two variations of Top2Vec were modeled, one with  Doc2Vec and one with the \textit{"all-mpnet-base-v2"} SBERT model\footnote{For an overview of SBERT models and their performance, see \url{https://www.sbert.net/docs/pretrained_models.html}}. 

For fair comparisons between BERTopic and Top2Vec, the parameters of HDBSCAN and UMAP were fixed between topic models. 

To measure the generalizability of BERTopic across language models, four different language models were used in the experiments with BERTopic, namely the Universal Sentence Encoder \cite{cer2018universal}, Doc2Vec, and the \textit{"all-MiniLM-L6-v2"} (MiniLM) and \textit{"all-mpnet-base-v2"} (MPNET) SBERT models. 

Finally, BERTopic, with and without the assumption of linearly-evolving topics, was compared with the original dynamic topic model, referred hereto as LDA Sequence.

\begin{table*}[ht]
    \centering
    \begin{tabular}{ccccccccc}
        \toprule
        {} & \multicolumn{2}{c}{\textbf{20 NewsGroups}} & \multicolumn{2}{c}{\textbf{BBC News}} 
        & \multicolumn{2}{c}{\textbf{Trump}}\\
        \cmidrule(lr){2-3} \cmidrule(lr){4-5} \cmidrule(lr){6-7} \cmidrule(lr){8-9}
        {} & TC & TD & TC & TD & TC & TD \\
        \midrule
        BERTopic-\textit{USE} &  .149 & .858 & .158 & .764 & .051 & .684\\
        BERTopic-\textit{Doc2Vec} & .173 & .871 & .168 & .819 & -.088\hspace{.11cm} & .536\\
        BERTopic-\textit{MiniLM} & .159 & .833 & .170 & .802 & .060 & .660\\
        BERTopic-\textit{MPNET} & .166 & .851 & .167 & .792 & .066 & .663\\
        \bottomrule
    \end{tabular}
    \caption{Using four different language models in BERTopic, coherence score (TC) and topic diversity (TD) were calculated ranging from 10 to 50 topics with steps of 10. All results were averaged across 3 runs for each step. Thus, each score is the average of 15 separate runs.}
    \label{tab:bertopic-lm}
\end{table*}

\subsection{Evaluation}
The performance of the topic models in this paper is reflected by two widely-used metrics, namely topic coherence and topic diversity. For each topic model, its topic coherence was evaluated using normalized pointwise mutual information (NPMI, \citep{bouma2009normalized}). This coherence measure has been shown to emulate human judgment with reasonable performance \citep{lau2014machine}. The measure ranges from [-1, 1] where 1 indicates a perfect association. Topic diversity, as defined by \cite{dieng2020topic}, is the percentage of unique words for all topics. The measure ranges from [0, 1] where 0 indicates redundant topics and 1 indicates more varied topics. 

Ranging from 10 to 50 topics with steps of 10, the NPMI score was calculated at each step for each topic model. All results were averaged across 3 runs for each step. To evaluate the dynamic topic models, the NPMI score was calculated at 50 topics for each timestep and then averaged. All results were averaged across 3 runs.  

Validation measures such are topic coherence and topic diversity are proxies of what is essentially a subjective evaluation. One user might judge the coherence and diversity of a topic differently from another user. As a result, although these measures can be used to get an indication of a model's performance, they are just that, an indication. 

It should be noted that although NPMI has been shown to correlate with human judgment, recent research states that this may only be the case for classical models and that this relationship might not exist with neural topic models \cite{hoyle2021automated}. In part, the authors suggest a needs-driven approach to evaluation as topic modeling's primary use is in computer-assisted content analysis. 

To this purpose, the differences in running times of each model were explored as they can greatly impact their usability. Here, we choose to focus on the wall times as it more accurately reflects how the topic modeling techniques would be used in practice. All the models are run on a machine with 2 cores of Intel(R) Xeon(R) CPU @ 2.00GHz and a Tesla P100-PCIE-16GB GPU. 

Moreover, in Section \ref{sec:discussion}, the strengths and weaknesses of the proposed model across use cases will be discussed extensively to further shed a light on what the model can and cannot do. 

\section{Results}
\label{sec:results}

Our main results can be found in Table \ref{tab:main-results}. 

\subsection{Performance}
From Table \ref{tab:main-results}, we can observe that BERTopic generally has high topic coherence scores across all datasets. It has the highest scores on the slightly preprocessed dataset, Trump's tweets, whilst remaining competitive on the thoroughly preprocessed datassets, 20 NewsGroups and BBC News. Although BERTopic demonstrates competitive topic diversity scores, it is consistently outperformed by CTM. This is consistent with their results indicating high topic diversity, albeit using a different topic diversity measure \cite{bianchi2020pre}. 

\subsection{Language Models}
The results in Table \ref{tab:bertopic-lm} demonstrate the stability of BERTopic, in terms of both topic coherence and topic diversity, across SBERT language models. As a result, the smaller and faster model, \textit{"all-MiniLM-L6-v2"}, might be preferable when limited GPU capacity is available. 

Although the USE and Doc2Vec language in BERTopic generally have similar performance, Doc2Vec scores low on the Trump dataset. This is reflected in the results we find in \ref{tab:main-results} where Top2Vec with Doc2Vec has poor performance. These results suggest that Doc2Vec struggles with creating accurate representations of the Trump dataset. 

On topic coherence, Top2Vec with Doc2Vec embeddings shows competitive performance. However, when MPNET embeddings are used both its topic coherence and diversity drop across all datasets suggesting that Top2Vec might not be best suited with embeddings outside of those generated through Doc2Vec. This is not unexpected as both word and documents vectors in Doc2Vec are jointly embedded in the same space, which does not hold for all language models. 

In turn, this also suggests why BERTopic remains competitive regardless of the embedding model. By separating the process of embedding documents and constructing the word-topic distribution, BERTopic is flexible in its embedding procedure. 

\subsection{Dynamic Topic Modeling}

From Table \ref{tab:dtm}, we can observe that BERTopic with and without the assumption of linearly evolving topics performs consistently well across both datasets. For Trump, it outperforms LDA on all measures whereas it only achieves the top score on topic coherence for the UN dataset. 

On both datasets, there seems to be no effect of the assumption of linearly evolving topics on both topic coherence and topic diversity indicating that from an evaluation perspective, the proposed assumption does not impact performance.

\begin{table}[ht]
    \begin{tabular}{lcccccccc}
        \toprule
        {} & \multicolumn{2}{c}{\textbf{Trump}} & \multicolumn{2}{c}{\textbf{UN}} \\
        \cmidrule(lr){2-3} \cmidrule(lr){4-5}
        {} & TC & TD & TC & TD \\
        \midrule
        LDA Sequence & .009 & .715 & .173 & \textbf{.820}\\
        BERTopic & \textbf{.079} & .862 & \textbf{.231} & .779\\
        BERTopic-\textit{Evolve} & \textbf{.079} & \textbf{.863} & .226 & .769\\
        \bottomrule
    \end{tabular}
    \caption{The topic coherence (TC) and topic diversity (TD) scores were calculated on dynamic topic modeling tasks. The TC and TD scores were calculated for each of the 9 timesteps in each dataset. Then, all results were averaged across 3 runs for each step. Thus, each score represents the average of 27 values.}
    \label{tab:dtm}
\end{table}

\begin{figure*}
  \includegraphics[width=\textwidth]{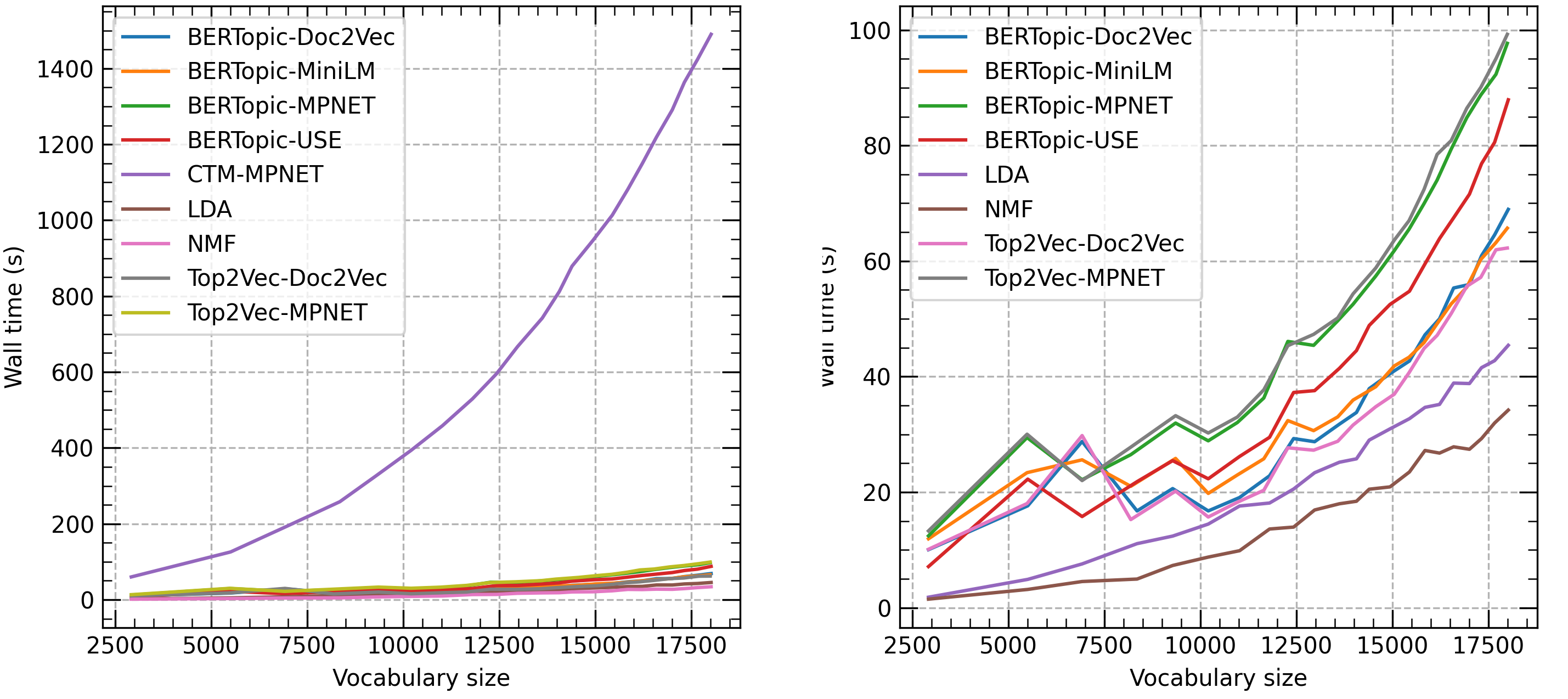}
  \caption{Computation time (wall time) in seconds of each topic model on the Trump dataset. Increasing sizes of vocabularies were regulated through selection of documents ranging from 1000 documents until 43000 documents with steps of 2000. \textbf{Left}: computational results with CTM. \textbf{Right}: computational results without CTM as it inflates the y-axis making differentiation between other topic models difficult to visualize.}
  \label{fig:run}
\end{figure*}

\subsection{Wall time}
From the left graph in Figure \ref{fig:run}, CTM using a MPNET SBERT model, is quite slow compared to all other models. If we remove that model from the results, we can more easily compare the wall time of the topic models that are more close in speed. Then, We can observe that the classical models, NMF and LDA, are faster than the neural network-based topic modeling techniques. Moreover, BERTopic and Top2Vec are quite similar in wall times if they are using the same language models. Interestingly, the MiniLM SBERT model seems to be similar in speed compared with Doc2Vec indicating that in BERTopic, MiniLM is a good trade-off between speed and performance. 

However, it should be noted that in the environment used in this experiment a GPU was available for creating the embeddings. As a result, the wall time is expected to increase significantly when embedding documents without a GPU. Although Doc2Vec can be used as a language model instead, previous experiments in this study have put its stability with respect to topic coherence and topic diversity into question.

\section{Discussion}
\label{sec:discussion}

Although we attempted to validate BERTopic across several experiments, topic modeling techniques can be validated through many other evaluation metrics, such as metrics for unsupervised and supervised modeling performance. Moreover, topic modeling techniques can be used across a variety of use cases, many of which are not covered in this study. For those reasons, we additionally discuss the strengths and weaknesses of BERTopic to further describe when, and perhaps most importantly, when not to use BERTopic.

\subsection{Strengths}
There are several notable strengths of BERTopic compared to the topic models used in this study. 

First, the experiments demonstrate that BERTopic remains competitive regardless of the language model used to embed the documents and that performance may increase when leveraging state-of-the-art language models. This indicates its ability to scale performance with new developments in the field of language models whilst still remaining competitive if classical language models are used. Moreover, its stability across language models allows it to be used in a wide range of situations. For example, when a user does not have access to a GPU, Doc2Vec can be used to generate competitive results. 

Second, separating the process of embedding documents from representing topics allows for significant flexibility in the usage and fine-tuning of BERTopic. Different preprocessing procedures can be used when embedding the documents and when generating the topic representations. For example, one might want to remove stopwords in the topic representations but not before creating document embeddings. Similarly, once the documents have been clustered, the topic generation process can be fine-tuned, by, for example, increasing the n-gram of words in the topic representation, without the need to re-cluster the data.  

Third, by leveraging a class-based version of TF-IDF, we can represent topics as a distribution of words. These distributions have allowed BERTopic to model the dynamic and evolutionary aspects of topics with little changes to the core algorithm. Similarly, with these distributions, we can also model the representations of topics across classes. 

\subsection{Weaknesses}
No model is perfect and BERTopic is definitely no exception. There are several weaknesses to the model that should be addressed. First, BERTopic assumes that each document only contains a single topic which does not reflect the reality that documents may contain multiple topics. Although documents can be split up into smaller segments, such as sentences and paragraphs, it is not an ideal representation. However, as HDBSCAN is a soft-clustering technique, we can use its probability matrix as a proxy of the distribution of topics in a document. This resolves the issue to some extent but it does not take into account that documents may contain multiple topics during the training of BERTopic. 

Second, although BERTopic allows for a contextual representation of documents through its transformer-based language models, the topic representation itself does not directly account for that as they are generated from bags-of-words. The words in a topic representation merely sketch the importance of words in a topic whilst those words are likely to be related. As a result, words in a topic might be similar to one another and can be redundant for the interpretation of the topic. In theory, this could be resolved by applying maximal marginal relevance to the top \textit{n} words in a topic but it was not explored in this study \cite{carbonell1998use}.

\section{Conclusion}
\label{sec:conclusion}
We developed BERTopic, a topic model that extends the cluster embedding approach by leveraging state-of-the-art language models and applying a class-based TF-IDF procedure for generating topic representations. By separating the process of clustering documents and generating topic representations, significant flexibility is introduced in the model allowing for ease of usability.  

We present in this paper an in-depth analysis of BERTopic, ranging from evaluation studies with classical topic coherence measures to analyses involving running times. Our experiments suggest that BERTopic learns coherent patterns of language and demonstrates competitive and stable performance across a variety of tasks.

\bibliography{anthology,custom}
\bibliographystyle{acl_natbib}

\end{document}